\documentclass[11pt,a4paper]{article}
\usepackage[hyperref]{PratyayaDerivativeNounAnalysis}
\usepackage{times}
\usepackage{latexsym}
\usepackage{graphicx}

\usepackage{microtype}

\aclfinalcopy % Uncomment this line for the final submission
\title{A Benchmark Corpus and Neural Approach for Sanskrit Derivative Nouns Analysis}

\author{Arun Kumar Singh \\
  Indian Institute of Technology - Delhi \\
  \texttt{mailtoarunkus@gmail.com} \\
  Sushant Dave \\
  Institute of Technology - Delhi \\
  \texttt{sushant.dave@gmail.com} \\
  Prof. Brejesh Lall \\
  Indian Institute of Technology - Delhi \\
  \texttt{brejesh@ee.iitd.ac.in} \\\And
    Dr. Prathosh A. P. \\
  Department of Electrical Engineering \\
  \texttt{prathoshap@ee.iitd.ac.in} \\
    Shresth Mehta \\
  Indian Institute of Technology - Delhi \\
  \texttt{shresthmehta2017@gmail.com} \\}

\date{}

\begin{document}
\maketitle
\begin{abstract}
This paper presents first benchmark corpus of Sanskrit Pratyaya (suffix) and inflectional words (padas) formed  due to suffixes along with neural network based approaches to process the formation and splitting of inflectional words. Inflectional words spans the primary and secondary derivative nouns as the scope of current work. Pratyayas are an important dimension of morphological analysis of Sanskrit texts. There have been Sanskrit Computational Linguistics tools for processing and analyzing Sanskrit texts. Unfortunately there has not been any work to standardize \& validate these tools specifically for derivative nouns analysis. In this work, we prepared a Sanskrit suffix benchmark corpus called Pratyaya-Kosh to evaluate the performance of tools. We also present our own neural approach for derivative nouns analysis while evaluating the same on most prominent Sanskrit Morphological Analysis tools. This benchmark will be freely dedicated and available to researchers worldwide and we hope it will motivate all to improve morphological analysis in Sanskrit Language.
\end{abstract}

\section{Introduction}
Sanskrit is considered as one of the oldest Indo-Aryan languages. The oldest known Sanskrit texts are estimated to be dated around 1500 BCE. A large corpus of religious, philosophical, socio-political and scientific texts of multi cultural Indian Subcontinent are in Sanskrit. Sanskrit, in its multiple variants and dialects, was the Lingua Franca of ancient India ~\citep{Harold:90}. Therefore, Sanskrit texts are an important resource of knowledge about ancient India and its people. Earliest known Sanskrit documents are available in the form called \emph{Vedic Sanskrit}. Rigveda, the oldest of the four Vedas, that are the principal religious texts of ancient India, is written in \emph{Vedic Sanskrit}. In sometime around 5\textsuperscript{th} BCE, a Sanskrit scholar named \emph{Panini} ~\citep{Cardona:97} wrote a treatise on Sanskrit grammar named \emph{Ashtadhyayi}, in which \emph{Panini} formalized rules on linguistics, syntax and grammar for Sanskrit. \emph{Panini}'s grammar is globally appreciated for its insightful analysis of Sanskrit and completeness of its descriptive coverage of the spoken standard language of \emph{Panini}'s time. 

\emph{Ashtadhyayi} \footnote{https://www.britannica.com/topic/Ashtadhyayi} is the oldest surviving text and the most comprehensive source of grammar on Sanskrit today and provides often unique information on Vedic, regional and  socio-linguistic usage. \emph{Ashtadhyayi} literally means eight chapters and these eight chapters contain around 4000 sutras or rules in total. These rules completely define the Sanskrit language as it is known today. \emph{Ashtadhyayi} is remarkable in its conciseness and contains highly systematic approach to grammar. Because of its well defined syntax and extensively well codified rules, many researchers have made attempts to codify the \emph{Panini}’s sutras as computer programs to analyze Sanskrit texts.
This paper tries to address the problem of unavailability of benchmark corpus and provides morphological analysis method for derivative nouns as a result of Sanskrit suffixes applied on root verbs and nouns using a machine learning approach.

\subsection{Introduction of Pratyaya in Sanskrit}
Different ways of inflectional word formation as mentioned by ~\citep{Kiparsky-Paul-2008} are as below:
\begin{quote}
\small
- [Root + Suffix] Root: desideratives, intensives.\\
- [Word + Suffix] Root: denominal verbs.\\
- [Root + Suffix] Stem: primary (\emph{krit}) suffixes.\\
- [Word + Suffix] Stem: secondary (\emph{taddhit}) suffixes.\\
- [Word + Word] Stem: compounding.\\
- [Root + Suffix] Word: verb inflection.\\
- [Stem + Suffix] Word: noun inflection.
\end{quote}

Here in this introduction, we will explain more about primary and secondary suffixes.
Sanskrit is a rich inflected language and depends on nominal and verbal inflections for communication of meaning ~\citep{Murali2014KRIDANTAAF}. A fully inflected unit is called pada. The \emph{subanta  padas} are the inflected nouns and the \emph{tinanta  padas} are the inflected verbs. 

\begin{figure*}[h]
  \centering
  \includegraphics[width=\linewidth]{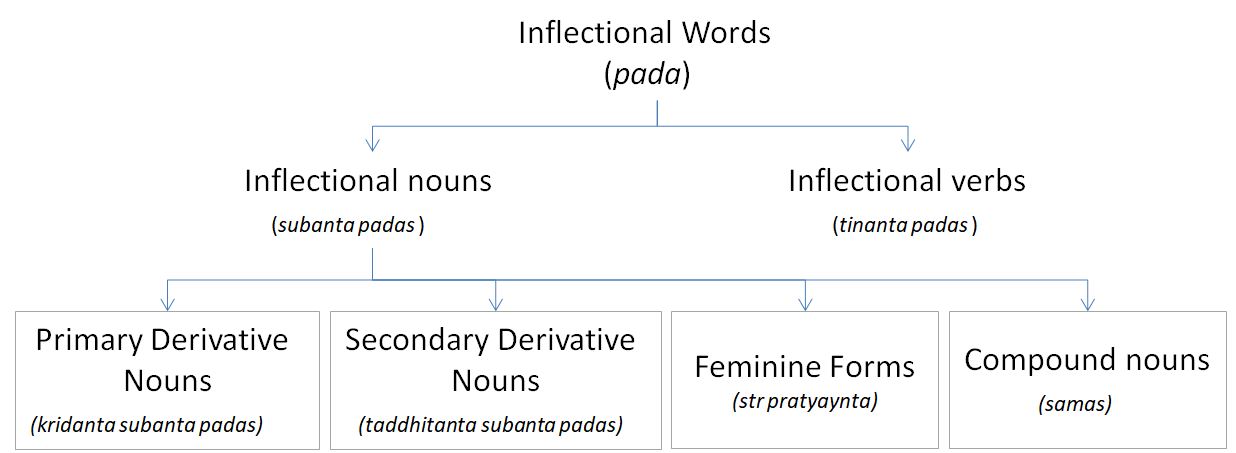}
    \caption{Inflectional Word Hierarchy}
  \label{fig:InflectionalWordHierarchy}
\end{figure*}

\subsubsection{Kridanta subanta (Primary Derivative Nouns)}
These are formed when the primary affixes called \emph{krit} are added to verbs to derive substantives, adjectives or indeclinable. \emph{DAtum nAma karoti iti krit. Kridanta} play a vital role in understanding Sanskrit language. Many morphological analyzers are lacking the complete
analysis of \emph{Kridanta}. Examples of \emph{krit} pratyaya are as below:
\begin{quote}
\small
\emph{paW} (root verb)+ \emph{tavya} (\emph{krit} suffix) = \emph{paWitavyam} \\
\emph{paW} (root verb)+ \emph{tumun}  (\emph{krit} suffix) = \emph{paWitum}
\end{quote}
\emph{krit} suffixes are mainly of seven types viz. \emph{tavyat, tavya, anIyara,Ryat, yat, kyap, kelimer}.

\subsubsection{Taddhitanta subanta (Secondary Derivative Nouns)}
The secondary derivative affixes called \emph{taddhit} derive secondary nouns from primary nouns. Some examples of \emph{taddhit} pratyaya are as below:
\begin{quote}
\small
\emph{Indra} (noun) + \emph{aR} (\emph{taddhit} suffix) = \emph{Endra} \\
\emph{Dana} (noun) + \emph{vatup} (\emph{taddhit} suffix) = \emph{Danavat}
\end{quote}
\emph{taddhit} suffixes are mainly of fourteen types.

\section{Existing Work on Sanskrit Derivative Nouns Analysis}
~\citet{chandra-2006} presented a model in the form of a tool for recognition and analysis of nominal morphology (Sanskrit \emph{subanta  padas}) in Sanskrit text for Machine Translation. The tool recognizes nominal morphology with the help of avyaya and verb database and does analysis of Sanskrit \emph{subanta  padas}. ~\citet{Murali2014KRIDANTAAF} provided an approach to deal with \emph{Kridanta}. Morphological dictionaries for upapadas, upasargas, roots and suffixes were created and rule based avyaya analyzer was developed which does Morphological Analysis based on identification \& filtering of upapadas, upsargas based on dictionary. 

The method adopted by ~\citet{Bharati-Akshar-2006} was a paradigm based approach where a student is taught the word forms of a common word e.g. deva in Sanskrit and that it is the default paradigm for 'a' ending masculine words. Further the list of exceptional words and the forms where they differ are taught separately. Following this method, a simple algorithm was developed which is described in ~\citep{Bharati-Akshar-2002}. This algorithm has been used to develop morph analyzer for different Indian languages (IIIT-H).
Separate modules have been developed to handle \emph{subanta, tinanta} and \emph{Kridanta} words. A list of lexicon is extracted from Monier William's dictionary. \emph{Kridanta} analyzer works based on rule based approach to retrieve pratyaya and upasarga form lexicon dictionary. It doesn’t provide any module to analyze other derivational suffixes such as \emph{taddhit} but claims for a provision of plug in.  ~\citet{Krishna2017AGB} proposed an approach for analysis of derivational nouns in Sanskrit. This approach attempted to build a semi supervised model that can identify usage of derived words in a corpus and map them to their corresponding source words. Special interest was given to the usage of secondary derivative affixes in Sanskrit ie. \emph{taddhit}.
When it comes to automating \emph{taddhit}, the Sanskrit Heritage System ~\citep{Goyal-Huet-2013} is an existing system that can recognize \emph{taddhit} and perform the analysis, but it does not generate the \emph{taddhit} and only the lexicalized \emph{taddhit} are recognized during the analysis. 

~\citet{Krishna2015TowardsAT} attempted to automate the process of deriving \emph{taddhit} in complete adherence to \emph{Ashtadhyayi}. The proposed system adopted a completely object oriented approach in modelling \emph{Ashtadhyayi}. The rules of \emph{Ashtadhyayi} were modeled as classes and were the environment that contains the entities for derivation. In their work the rule group was achieved through formation of inheritance network.
The current resources available for finding out derivational nouns in open domain are not very accurate. Two most popular publicly available set of Pratyaya Analysis tools viz. JNU Sanskrit \emph{Kridanta} Analyzer \footnote{http://sanskrit.jnu.ac.in/kridanta/ktag.jsp} and UoH Morphological Generator \footnote{http://sanskrit.uohyd.ac.in/scl/} are mentioned in table ~\ref{table:Open Available Tools for Derivative Nouns (Pada) Analysis}.

\begin{table*}
\centering
\begin{tabular}{|p{4cm} p{10cm}|}
\hline
\textbf{Tool Name} & \textbf{Description}\\
\hline
Sanskrit \emph{Kridanta} Analyzer (JNU) & The "Sanskrit \emph{Kridanta} Analyzer" was partially completed as part of M.Phil. research  submitted to Special Center for Sanskrit Studies, JNU in 2008 by Surjit Kumar Singh  (M.Phil 2006-2008) under the supervision of Dr. Girish Nath Jha . It facilitates the split of \emph{Kridanta} into root verb and \emph{krit} pratyaya \\
Morphological Generator (UoH)
 & Morphological Generator shows the inflectional, and (some) derivational forms of a given noun or a verb. This tool has been developed by University of Hyderabad (UoH)  \\\hline
\end{tabular}

\caption{Open Available Tools for Derivative Nouns (Pada) Analysis}
\label{table:Open Available Tools for Derivative Nouns (Pada) Analysis}
\end{table*}

\section{Motivation}
Researchers have earlier attempted to develop morphological analyzer for Sanskrit such as cdac \footnote{http://www.cdac.in/html/ihg/ihgidx.asp}, Sanskrit academy  \footnote{http://www.sanskritacademy.org/} and ~\citep{Huet2003TowardsCP}). But having constraint of limited coverage or not being available openly, it is difficult to reuse it for further applications in NLP area. Also it is not possible for a person from a non-Sanskrit background to develop applications and systems in Sanskrit morphology, even though Sanskrit is well codified language. Among various facets of Morphological Analysis, suffix analyzer is necessary for a good coverage morphological analyzer. Our proposed Sanskrit noun derivatives analyzer will facilitate the work in below areas.

\begin{itemize}
\small
\itemsep0em
\item Text to speech synthesis system
\item Neural Machine Translation from non-Sanskrit to Sanskrit language and vice versa
\item Sanskrit Language morphological analysis 
\end{itemize}

Pratyayas are not easily available for building Sanskrit based applications either in printed or e-forms. Though some sources are available such as \emph{Panini Pratyayartha Kosha Taddhita Prakaranam} and \emph{Panini Kridanta Pratyaya Artha Kosha} by Dr. Gyanprakash Shastri   \footnote{https://archive.org/} , \emph{Laghu Siddhant Kaumudi}  \footnote{https://chaukhambapustak.com/} and \emph{Kridant Karak Prakaranam} by Bhimsen Shashtri but one has to have good understanding of Sanskrit to analyze it and use it to develop some analysis tool or system. We have prepared Pratyaya-Kosh in the form of root, suffix and derivative noun due to affix. This data can be easily used to develop analyzers and can be trained to learn rules using various algorithms automatically.

\section{Method}
\subsection{Kridanta Pada Formation Method}
\label{sec:Kridanta Pada Formation}
\emph{Kridanta} Pada formation task is similar to language translation problem where a sequence of characters or words produces another sequence and lengths of the inputs and outputs are not fixed. RNNs are widely used to solve such problems. Sequence to sequence model introduced by ~\citep{sutskever2014sequence} is especially suited to such problems, therefore it was used in this work. The training and test data were in ITRANS Devanagari format  \footnote{https://www.aczoom.com/itrans/}. This data was converted to SLP1  \footnote{https://en.wikipedia.org/wiki/SLP1} as SLP1 was found more suited for proposed approach. The code was implemented in python 3.5 with Keras API running on TensorFlow backend. 

\begin{figure*}[h]
  \centering
  \includegraphics[width=\linewidth]{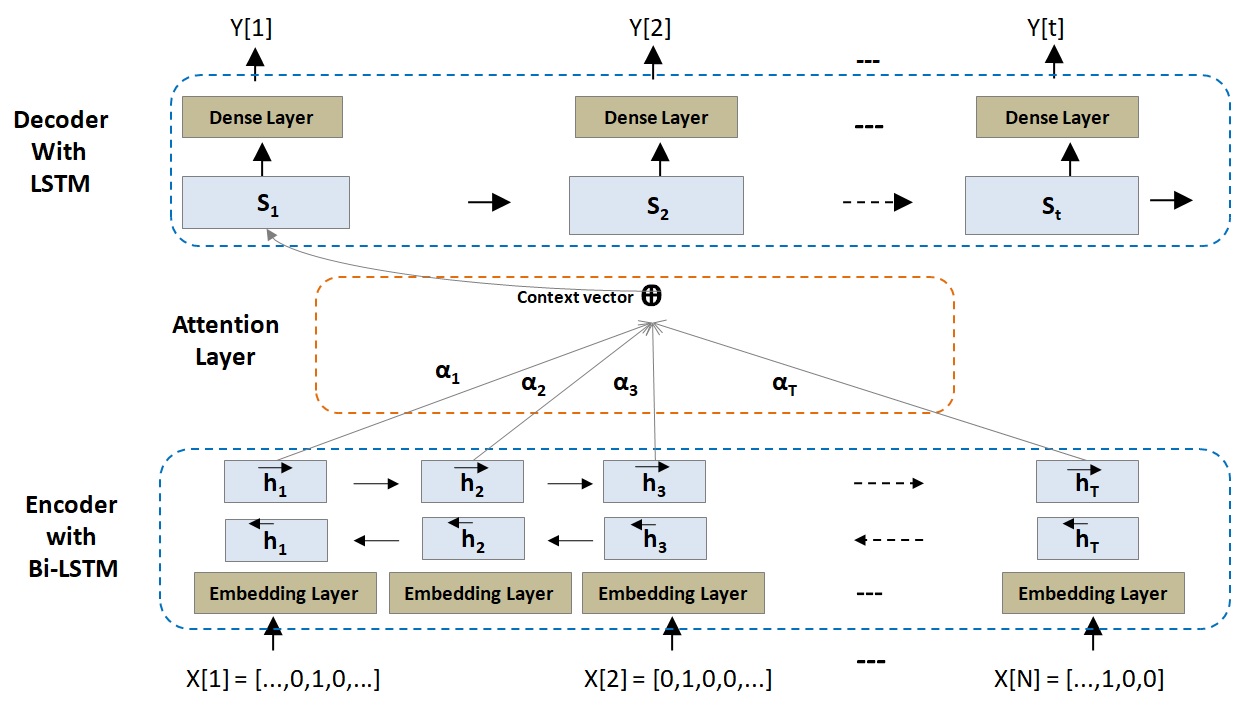}
    \caption{Architecture for Learning Derivative Noun (Pada) Formation }
  \label{fig:Pratyaya Model}
\end{figure*}

Proposed model expects 2 words (root + \emph{krit} pratyaya) concatenated using a "+" symbol as input and outputs the primary derivative noun (\emph{Kridanta}) as shown in the example below:
\begin{quote}
\small
\emph{tul} + \emph{lyuw} = \emph{tolanam}
\end{quote}

Results achieved from this approach seem to be good. Proposed approach tries to address this problem by giving all the characters of the first input word and all the characters of the second input word concatenated with a ‘+’ symbol as the input to the encoder LSTM of the sequence to sequence model. Due to unavailability of enough data we based our analysis on the 10 pratyayas out of 12.
Input sequence is set as the two input words concatenated with a ‘+’ character between the 2 words. Output sequence is the \emph{Kridanta} (primary derivative noun) single word. Characters ‘\&’ and ‘\$’ were used as start and end markers respectively in the output sequence. The maximum lengths of input and output sequences were 17 and 18 respectively. So we used ‘*’ for padding shorter input sequences. We used the same dictionary for input and output characters. The dictionary contained just 53 characters and characters usually don't have too much correlation as can be observed for words. This is the reason that we used one hot encoded representation of the characters instead of training any word embedding. The best results were achieved with an LSTM ~\citep{hochreiter1997long} as basic RNN cell for decoder and bidirectional LSTM as basic RNN cell for encoder. Adding attention to the encoder part improved the accuracy slightly so we opted for using attention in our model. Both the encoder and decoder use the hidden unit size (latent dimension) as 128. The training vectors were divided in batches of 32 vectors and total of 70 epochs were run to get the final results.

\subsection{Taddhitanta Pada Formation Method}
\emph{Taddhitanta} are formed from \emph{taddhit} pratyayas much like \emph{Kridanta} are formed from \emph{krit} pratyayas. Below example shows this formation:
\begin{quote}
\small
\emph{sUrasena} + \emph{Rya} = \emph{sOrasenyaH}
\end{quote}

This is similar to \emph{Kridanta} formation and thus we used the same sequence to sequence model.
However the data that we had in case of \emph{taddhit} pratyaya was limited and imbalanced for training the neural network. The rules for \emph{Taddhitanta} formation moreover are considered to be more complicated than those for \emph{Kridanta} formation. We believe these are the reasons we did not get very high but a satisfactory accuracy on \emph{Taddhitanta} unlike \emph{Kridanta} which performed much better. The accuracy mentioned is calculated as:
\begin{quote}
\small
(Number of correct formations) / (Total number of test samples)
\end{quote}

The character wise accuracy, however is quite high (~98\%), which means that the model incorrectly predicts just one or two characters rather than whole words in incorrect formations.

\subsection{Derivative Noun (Pada) Split Method}
Conceptually, the model architecture explained in Section ~\ref{sec:Kridanta Pada Formation} for \emph{Kridanta} formation and \emph{Taddhitanta} formation respectively, can also be used for splitting the \emph{Kridanta} and \emph{Taddhitanta} back into root and pratyaya, where the input and output are swapped with compound word as input and the two initial words concatenated with ‘+’ character as output. The accuracy achieved with this approach was better than that obtained for pada formation. Our model seems to perform fairly well even for longer sequences of length greater than 15.

\section{Data and Evaluation Results}
\subsection{Pratyaya-Kosh}
Data for Pratyaya-Kosh was extracted from 4 different sources viz. Panini Pratyayarth kosh, Sanskrit Hindi Kosh dictionary, KridantaRupaMala \& Sankrit Abhyas Portal developed based on rules of \emph{Ashtadhyayi}. Majority of data was extracted from Sanskrit Abhyas Portal. Table ~\ref{table:Pratyaya-Kosh-Source} shows details of data sources.

\begin{table*}
\centering
\begin{tabular}{|p{4cm} p{4cm} p{6cm}|}
\hline
\textbf{Source Name} & \textbf{Author} & \textbf{Source Reference}\\
\hline
Sankrit Abhyas Portal & Sharat Kotian & http://www.sanskritabhyas.in/en \\
Panini Pratyayarth kosh & Dr. Gyanprakash Shastri & https://archive.org/details/PaniniPratyay 
arthaKoshaTaddhitaPrakaranamDr.GyanprakashShastri \\
KridantaRupaMala & Pandit S. Ramasubba Sastri & https://archive.org/details/
KridantaRupaMalaVol.1-5\\
Sanskrit Hindi Kosh dictionary & V.S. Apte & https://archive.org/details/
SanskritHindiKoshV.S.Apte\\\hline
\end{tabular}
\caption{Pratyaya-Kosh Data Sources}
\label{table:Pratyaya-Kosh-Source}
\end{table*}

This Pratyaya-Kosh corpus consists of \emph{Kridanta} (Primary derivative nouns) and \emph{Taddhitanta} (Secondary Derivative nouns) in a form which can be directly ingested for machine learning based approaches. Each record is a tuple in the form of (verb stem, \emph{krit} suffix, \emph{Kridanta}) and (noun, \emph{taddhit} suffix, \emph{Taddhitanta}).
Corpus size of \emph{Kridanta} padas is 24,757 which includes padas formed by 12 different \emph{krit} suffixes, where as corpus size of \emph{Taddhitanta} padas is 3,088 which are formed by 17 different \emph{taddhit} suffixes. Details of the corpus are given in table ~\ref{table:Pratyaya-Kosh-Details}.

\begin{table*}
\centering
\begin{tabular}{|p{3cm} p{3cm} p{3cm} p{3cm}|}
\hline
\textbf{\emph{krit} Pratyaya} & \textbf{Corpus Size} & \textbf{\emph{taddhit} Pratyaya} & \textbf{Corpus Size}\\
\hline
\emph{lyuw} & 2198 & \emph{Ca} & 152\\
\emph{anIyar} & 2198 & \emph{QaY} & 58\\
\emph{Rvul} & 2198 & \emph{Qak} & 124\\
\emph{tumu~n} & 2198 & \emph{Rya} & 104\\
\emph{tavya} & 2198 & \emph{Rini} & 36\\
\emph{tfc} & 2198 & \emph{WaY} & 150\\
\emph{ktvA} & 2198 & \emph{Wak} & 308\\
\emph{Lyap} & 2198 & \emph{aR} & 610\\
\emph{ktavatu~} & 2198 & \emph{aY} & 246\\
\emph{Kta} & 2198 & \emph{iY} & 96\\
\emph{Satf~} & 1687 & \emph{Itac} & 92\\
\emph{SAnac} & 1090 & \emph{Matup} & 168\\
- & - & \emph{Mayaw} & 10\\
- & - & \emph{Tal} & 312\\
- & - & \emph{Tva} & 311\\
- & - & \emph{yaY} & 113\\
- & - & \emph{Yat}  & 198\\\hline

\end{tabular}
\caption{Pratyaya-Kosh Corpus Details}
\label{table:Pratyaya-Kosh-Details}
\end{table*}

\subsection{Derivative Noun (Pada) Analysis Evaluation}
Training and test data was taken from Pratyaya-Kosh. In case of \emph{Kridanta} Padas, training data was chosen for 10 \emph{krit} suffixes out of 12 due to data unbalancing issue. \emph{Satf~} \& \emph{SAnac} suffixes have less data as compared to other \emph{krit} suffixes, hence \emph{Kridanta} Padas were left out corresponding to these 2 suffixes while training and testing. In case of \emph{Taddhitanta} padas analysis, all suffixes were taken into consideration for training and testing.
Analysis of padas (primary derivative nouns \& secondary derivative nouns) were done in two ways as pada formation and pada split into its original verb stem \& suffix in case of \emph{Kridanta} pada and noun \& suffix in case of \emph{Taddhitanta} pada. For analysis of both the cases, pada formation and pada split, total records (tuples) were considered as 21,980 for \emph{Kridanta} and 3088 for \emph{Taddhitanta}. 80\% examples used for training \& validation ( 17,584 for \emph{Kridanta}, 2470 for \emph{Taddhitanta}) and remaining 20\% examples used for model testing(4396 for \emph{Kridanta} and 618 for \emph{Taddhitanta}). 
In case of Derivative noun or pada formation, evaluation is based on exact match of whole pada. To evaluate the pada split, both suffix as well as original root stem or noun should be correct to consider it as success.  Even if the model confuses between a \& A (SLP1 encoding), it is considered a failure. 
Results from method described above were bench marked with results from other publicly available tools as mentioned in table ~\ref{table:Open Available Tools for Derivative Nouns (Pada) Analysis}. Test set from Pratyaya-Kosh was kept same for comparing with UoH \& JNU tools. UoH tool does pada split for  both primary as well as secondary derivative nouns where as JNU tool does pada split for only primary derivative nouns. However both these tools does not provide feature for pada formation analysis. Evaluation criteria was kept strict for proposed method as mentioned above while the success criteria for UoH \& JNU tool was relaxed in a way if either of original stem verb/noun or suffix is found in the result, it is considered as success.

The comparison is shown in the table ~\ref{table:Benchmark-Results-Derivative-noun-split}. Every cell in the table ~\ref{table:Benchmark-Results-Derivative-noun-formation}  \&~\ref{table:Benchmark-Results-Derivative-noun-split} indicates successful test cases, overall test cases and success percentage.

\begin{table*}
\centering
\begin{tabular}{|p{4cm} p{4cm} p{4cm}|}
\hline
\textbf{Model} & \textbf{\emph{Kridanta} Formation Accuracy} & \textbf{\emph{Taddhitanta} Formation Accuracy}\\
\hline
JNU (Sanskrit \emph{Kridanta} Analyzer) & - & - \\
UoH (Morphological Generator) & - & - \\
Proposed Method & 3718 / 4396 (84.58 \%) & 495 / 618 (80.09\%)\\\hline
\end{tabular}
\caption{Benchmark Results of Derivative noun (Pada) formation}
\label{table:Benchmark-Results-Derivative-noun-formation}
\end{table*}

\begin{table*}
\centering
\begin{tabular}{|p{4cm} p{4cm} p{4cm}|}
\hline
\textbf{Model} & \textbf{\emph{Kridanta} Split Prediction Accuracy} & \textbf{\emph{Taddhitanta} Split Prediction Accuracy}\\
\hline
JNU (Sanskrit \emph{Kridanta} Analyzer) & 1710 / 4396
(38.9\%)
 & - \\
UoH (Morphological Generator) & 3492 / 4396
(79.43\%)
 & 146 / 618
(23.62\%)
 \\
Proposed Method & 3903 / 4396
(88.79 \%)
 & 255 / 618
(41.26\%)
\\\hline
\end{tabular}
\caption{Benchmark Results of Derivative noun (Pada) Split}
\label{table:Benchmark-Results-Derivative-noun-split}
\end{table*}

It is evident form results that proposed method improves upon the existing Morphological Analysis tools and methods by a significant margin. In addition, proposed model does not require any external lexicon for analysis and its prediction mechanism works better than dictionary based tools and approaches.

\section{Conclusion}
In this research work, we propose Pratyaya-Kosh, a benchmark corpus to help researchers new to Sanskrit in building AI based Morphological Analyzer for Sanskrit derivative nouns. Also we propose neural approach for learning derivative noun formation without use of any external resources such as language models, morphological or phonetic analyzers and still manage to outperform existing approaches. In future we intend to extend current work to verb derivative and indeclinable derivative using machine learning methods. Proposed models can be further refined by using additional training data. Benchmark corpus (Pratyaya-Kosh) will be made available on git hub.
    
%\section*{Acknowledgments}
%The authors would like to thank Mr. Sharat Kotian for his work in codifying sutras of %\emph{Ashtadhyayi} which helped in constructing the Corpus for primary (\emph{Kridanta}) and secondary %derivatives (\emph{Taddhitanta}) nouns.

\bibliography{PratyayaDerivativeNounAnalysis}
\bibliographystyle{acl_natbib}

\end{document}